\documentclass[10pt, a4paper]{article}

\usepackage{lrec-coling2024} 
\usepackage{latexsym}
\usepackage{microtype}
\usepackage{booktabs}
\usepackage{multirow}
\usepackage{graphicx}
\usepackage{amssymb}
\usepackage{amsmath}
\usepackage{pifont}
\usepackage{colortbl}
\usepackage{caption}
\usepackage{subcaption}
\usepackage{tikz}
\usepackage{color}
\usepackage{enumitem}
\usepackage{cleveref}

\setlist[itemize]{noitemsep, nolistsep, leftmargin=*}
\setlist[enumerate]{noitemsep, nolistsep, leftmargin=*}

\newcommand\tikzmark[1]{\tikz[overlay,remember picture] \node (#1) {};}

\title{Improving Language Model Reasoning with Self-motivated Learning}

\name{Yunlong Feng$^{\dagger}$, Yang Xu$^{\dagger}$, Libo Qin$^{\dagger}$, Yasheng Wang$^{\ddag}$, Wanxiang Che$^{\dagger}$\sthanks{~~Corresponding author.}}

\address{$^{\dagger}$Research Center for Social Computing and Information Retrieval \\
Harbin Institute of Technology, China \\
{\{ylfeng, yxu, lbqin, car\}@ir.hit.edu.cn} \\
$^{\ddag}$Huawei Noah's Ark Lab  \\
yashengwang@huawei.com}

\abstract{
  Large-scale high-quality training data is important for improving the performance of models. 
  After trained with data that has rationales (reasoning steps), models gain reasoning capability.
  However, the dataset with high-quality rationales is relatively scarce due to the high annotation cost.
  To address this issue, we propose \textit{Self-motivated Learning} framework. 
  The framework motivates the model itself to automatically generate rationales on existing datasets.
  Based on the inherent rank from correctness across multiple rationales, the model learns to generate better rationales, leading to higher reasoning capability.
  Specifically, we train a reward model with the rank to evaluate the quality of rationales, and improve the performance of reasoning through reinforcement learning.
  Experiment results of Llama2 7B on multiple reasoning datasets show that our method significantly improves the reasoning ability of models, even outperforming text-davinci-002 in some datasets.
 \\ \newline 
}

\begin{document}

\maketitleabstract

\section{Introduction}\label{sec:intro}

Large Language Models (LLMs) that are pre-trained on extensive text corpora have exhibited profound capability across a diverse array of downstream tasks. Particularly, their adaptability in both few-shot and zero-shot learning contexts, achieved by assimilating task-specific instructions and demonstrations, has garnered significant attention \citep{raffel2020exploring,brown2020language,zhang2022opt,chowdhery2022palm,lampinen2022can,gu-etal-2022-learning,ye2023context}.
This approach emphasizes generating a series of intermediate reasoning steps, which can be achieved through CoT demonstrations in prompts \cite{wei2022chain} or by guiding models with instructions in zero-shot scenarios \cite{NEURIPS2022_8bb0d291}.

Some studies have demonstrated that large-scale, high-quality data is crucial for enhancing the reasoning abilities of models \cite{kim2023cot, ho-etal-2023-large, strategyqa__geva2021did,cobbe2021training}. But there is a scarcity of datasets with reasoning steps due to the high annotation cost.
On one hand, series of works resort to manual annotations for datasets \cite{lu2022learn, xie-etal-2020-worldtree, Mihaylov2018CanAS, khot2020qasc}. Training models with data obtained in this manner can significantly enhance their performance, albeit at a substantial cost.
On the other hand, some studies generate data using large-scale models \cite{ho-etal-2023-large, kim2023cot, luo2023wizardmath, liu2023visual,wang-etal-2023-scott,li-etal-2023-symbolic}, utilizing such data to train models to improve their performance. Both manual annotation and generating data using large models incur considerable costs.
Beyond these methods, some strategies involve models generating rationales and filtered by answers, subsequently using this data for finetuning \cite{zelikman2022star}.
In contrast, we propose a method that motivates the model itself to generate rationales of varying quality with existing datasets, integrating this preference into reinforcement learning to improve model performance.

\begin{figure}[t]
  \centering
  \includegraphics[page=1,width=\linewidth]{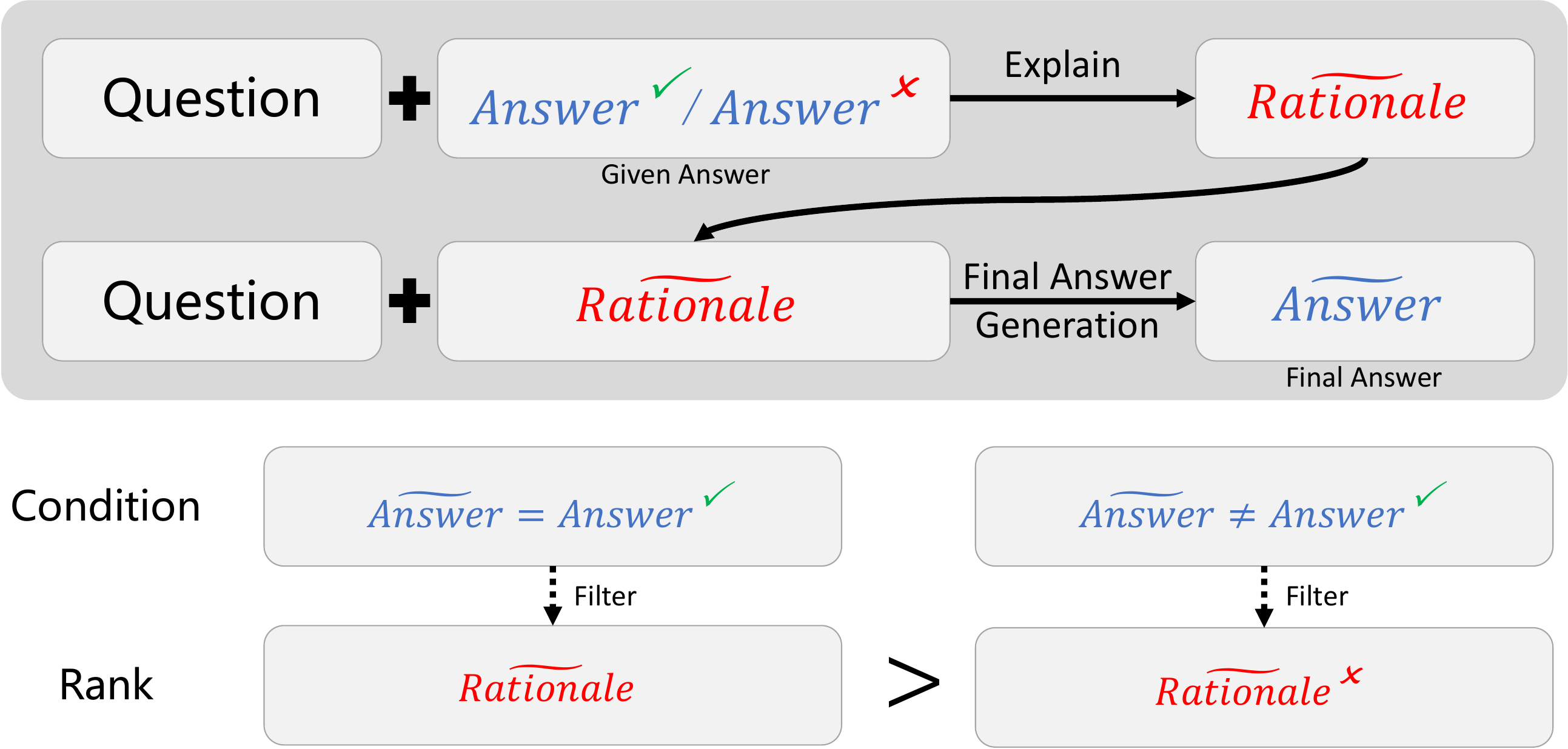}
  \caption{The motivation of our method. We note that the $\widetilde{Rationale}$ and $\widetilde{Answer}$ means they are generated by the language model. The main idea is that: (1) The correct given answer more likely leads to correct rationale. (2) The rationale that leads to the correct answer is better than the rationale that leads to the wrong answer.}
  \label{fig:overview}
\end{figure}

\begin{figure*}[ht]
  \centering
  \includegraphics[page=2,width=\textwidth]{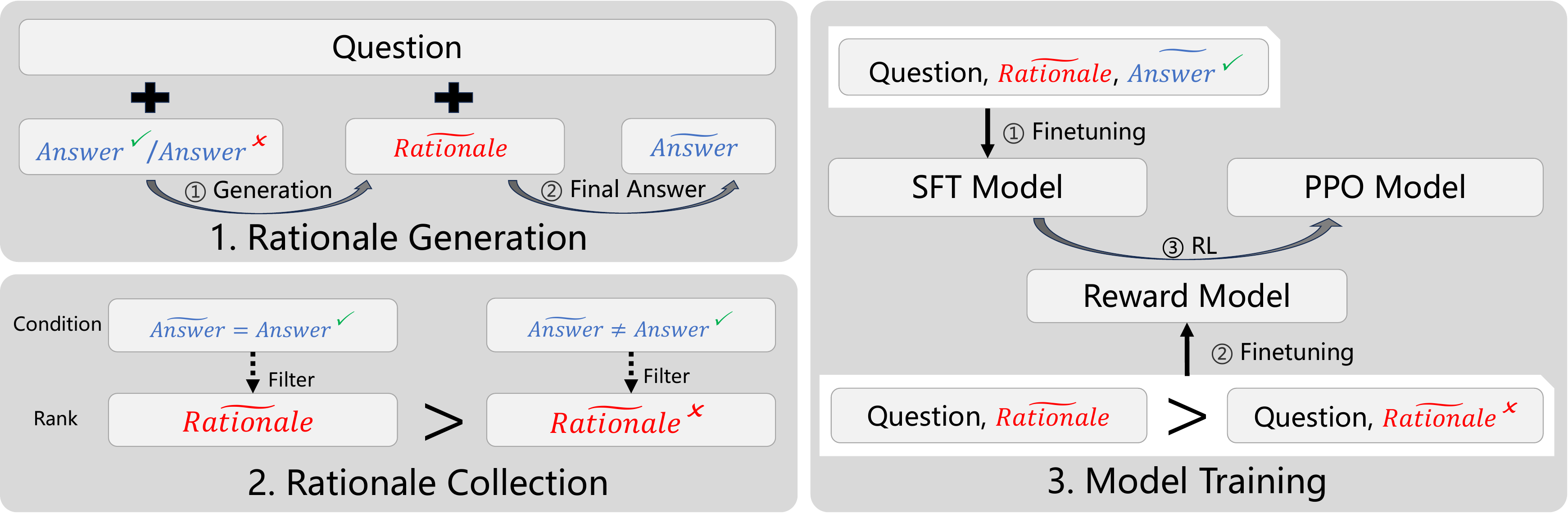}
  \caption{Overview of our method. As shown in the figure, our method can be divided into three steps in general: (1) \textbf{Rationale Generation}: We first generate a rationale using \textit{Few-shot-CoT} \cite{NEURIPS2022_8bb0d291}. Specifically, we first generate rationales for different answers (called \textit{Given Answer}). We then use these generated rationales to generate the \textit{Final Answer}. This allows us to later filter the rationales using both the original and generated answers.
    (2) \textbf{Rationale Collection}: We filter the rationales by determining whether the \textit{Given Answer} and the \textit{Final Answer} match the correct answer. This helps us identify relatively better and worse rationales.
    (3) \textbf{Model Traning}: We use the better data to train a model and get a \textit{Supervised Fine-tuning Model (SFT Model)}. We then use data of varying quality to build a \textit{Reward Model (RM)}. Finally, we utilize the previously acquired \textit{SFT Model} and \textit{Reward Model} for reinforcement learning with PPO.}
  \label{fig:method}
\end{figure*}

We ask the question \textit{Can we leverage the intrinsic properties of model-generated data to address the issue of data scarcity?}
We observe that there exists an inherent preference here, that a rationale capable of generating the correct answer should be superior to a rationale that generates an incorrect one. The implicit information in this statement is that proper reasoning should lead to correct results, while improper reasoning should yield incorrect ones.
To illustrate, consider the question ``Sam had 9 dimes in his bank. His father gave him 7 more. How many dimes does Sam possess now?'' The rationale ``Sam has 9 + 7 = 16 dimes.'' results in the correct answer ``16'', while the rationale ``Sam had 9 - 7 = 2 dimes remain.'' produces an incorrect answer ``2''. It is evident that the former rationale should be better than the latter.

This preference information can be used to filter or evaluate the quality of rationale. Specifically, we can use a model to generate a large number of rationales, and then utilize this preference to filter them, obtaining relatively high-quality and low-quality rationales. Subsequently, we can use this preference information to train a reward model that assesses the quality of rationales generated by the model.
By integrating current reinforcement learning algorithms, this reward model can be used to optimize the model, enabling it to discern which rationales are superior, thereby enhancing its performance. Through this approach, we can leverage existing datasets to generate rationales and construct a rank, solely relying on the model itself, without the need for extrenal large models or manual annotations. We call it Self-motivated Learning.

In summary, our contributions are as follows:
\begin{itemize}
  \item We point out an inherent preference in rationales, that is, a \textit{rationale} capable of generating correct answers should be superior to a \textit{rationale} generating incorrect answers. This preference reflects the quality of the \textit{rationale}.
  \item By using this preference, we alleviate data scarcity. We utilize the model and existing datasets to generate \textit{rationale}, integrating this preference into reinforcement learning to improve model performance.
  \item We conducted experiments using \textit{Llama 2 7B} and multiple datasets. The results demonstrate that our method can significantly improve the performance of the model. Without resorting to large models, our approach surpasses the performance of models fine-tuned with \textit{rationales} generated by \textit{text-davinci-002}. In some tasks, it even outperforms \textit{text-davinci-002}.
\end{itemize}

\section{Method}\label{sec:method}

\begin{table}[t]
  \centering
  \resizebox{\linewidth}{!}{%
    \begin{tabular}{p{0.6\linewidth}|p{0.6\linewidth}}
      \toprule
      Rationale    Generation                                            & Final Answer     Generation                        \\ \midrule
      {[}Instruction and Question{]}                                     & {[}Instruction and Question{]}                     \\
      <$q_i$>                                                            & <$q_i$>                                            \\
      {[}Answer{]}                                                       & {[}Rationale{]}                                    \\
      <$a'$>                                                             & \tikzmark{end}<$\hat{r}_i$>                        \\
      {[}Rationale{]}                                                    & {[}Answer{]}                                       \\
      \textcolor{white!50!black}{\textit{<$\hat{r}_i$>}}\tikzmark{start} & \textcolor{white!50!black}{\textit{<$\hat{a}_i$>}} \\ \bottomrule
    \end{tabular}%
    \begin{tikzpicture}[overlay, remember picture]
      \draw[->, thick, white!50!black] (start.east) to[out=0, in=180] (end.west);
    \end{tikzpicture}
  }
  \caption{Prompt templates for generation. We first use the question <$q_i$> and given answer <$a'$> to generate the rationale \textcolor{white!50!black}{\textit{<$\hat{r}_i$>}}, and then use the question <$q_i$> and generated rationale \textit{<$\hat{r}_i$>} to generate the final answer \textcolor{white!50!black}{\textit{<$\hat{a}_i$>}}.}
  \label{tab:templates}
\end{table}

We propose Self-motivated Learning, a task-agnostic approach to further improve the performance of reasoning in LMs.
The core idea is to generate correct/incorrect rationales with in-context learning, and then use them to do fine-tuning and reinforcement learning.
To filter the generated rationales, we use the model to generate the answer based on the generated rationales and compare it with the given answer and ground truth.
After filtering, we can get relatively high-quality and low-quality rationales to train the models.

\subsection{Training Process}\label{sec:training}

\paragraph{Step 1. Rationale Generation.}

As \cref{tab:templates} shows, we first utilize a model to generate rationales for a given task $\mathcal{T}$. Consider a standard sample $S_i$ consisting of a question $q_i$ and its true answer $a_i$. Using Few-shot-CoT \cite{NEURIPS2022_8bb0d291}, we prompt the model to generate a rationale $\hat{r}_i$ based on the given answer $a_i'$, where $a_i'$ is the correct answer or incorrect answer. Then we utilize a model to generate a final answer $\hat{a}_i$ for $\hat{r}_i$ with greedy decoding.
We can construct correct answers or incorrect answers from the source dataset. For example, in the question ``Would a pear sink in water?'', the correct answer is ``No''.
When generating a rationale, we place the given answer before the rationale, as \cref{tab:prompt} shown.
Besides the correct answer provided by the dataset, we can generate incorrect answers based on the given options or randomly.
In this way, we can produce rationales for both the correct and incorrect answers.
This provides some information for our subsequent rationale selection.

\begin{table}[t]
  \centering
  \resizebox{\linewidth}{!}{%
    \begin{tabular}{p{0.6\linewidth}|p{0.6\linewidth}}
      \toprule
      Prompt For Better Rationale                                                                                                                                            & Prompt For Worse Rationale                                                                                                                                            \\ \midrule
      {[}Instruction and Question{]}                                                                                                                                         & {[}Instruction and Question{]}                                                                                                                                        \\
      Would a pear sink in water?                                                                                                                                            & Would a pear sink in water?                                                                                                                                           \\
      {[}Answer{]}                                                                                                                                                           & {[}Answer{]}                                                                                                                                                          \\
      \textcolor{green!60!black}{No}                                                                                                                                         & \textcolor{red!80!black}{Yes}                                                                                                                                         \\
      {[}Rationale{]}                                                                                                                                                        & {[}Rationale{]}                                                                                                                                                       \\
      \textcolor{white!50!black}{\textit{The density of a pear is about 0.6g/cm3, which is less than water. Objects less dense than water float. Thus, a pear would float.}} & \textcolor{white!50!black}{\textit{The density of a pear is about 0.6g/cm3, which is less than water. Objects less dense than water float. Thus, a pear would sink.}} \\
      \bottomrule
    \end{tabular}%
  }
  \caption{Example of the prompt for better and worse rationale. The correct answer of the question is ``\textcolor{green!60!black}{No}''. We use the correct answer and incorrect answer to generate the \textcolor{white!50!black}{\textit{rationales}}.}
  \label{tab:prompt}
\end{table}

\paragraph{Step 2: Rationale Collection.}
Once we've generated the CoT rationales and the final answers, our next step is to filter these rationales based on their quality. Our objective is to distinguish between high-quality and low-quality rationales. We employ the following filtering criteria:

\begin{itemize}
  \item \textbf{Answer Consistency Check:} Evaluate the correctness of the provided answer \(a_i'\) and the final answer \(\hat{a}_i\) by comparing them with the true answer \(a_i\). When both \(a_i'\) and \(\hat{a}_i\) are correct, we categorize the corresponding rationale as a high-quality rationale. Conversely, if both are incorrect, the rationale is deemed low-quality. We throw away the rationales that do not fall into either of these categories.

  \item \textbf{Rationale Content Check:} We filter rationales that include the correct answer but exclude incorrect answers as high-quality rationales. In contrast, we discard the rationales that do not contain the correct answer as low-quality rationales.

  \item \textbf{Label Reference Check:} When dealing with multiple-choice questions, the given rationale should reference the label of the chosen answer. So, if ``C. Paris'' is the selected option, the word ``Paris'' should be incorporated in the rationale content.

  \item \textbf{Numerical Accuracy Check:} For numerical solutions, the answers are transformed into a floating-point format for consistency. If the absolute difference between two answers is less than \(1e-6\), they are treated as identical. Moreover, the answer should be present within the rationale.
\end{itemize}

\paragraph{Step 3. Model Training.}

After the generation and filtration processes in the initial two steps, we have obtained rationales of both relatively high and low quality. Then we can train the models with them.

\begin{table*}[ht]
  \resizebox{\linewidth}{!}{
    \begin{tabular}{lccclll}
      \toprule
      Dataset            & Choices & Training Samples & Test Samples & Data Split & License     & References                                           \\
      \midrule
      SingleEq           & -       & 356              & 152          & 70:30      & None        & \citetlanguageresource{singleeq__koncel2015parsing}  \\
      AddSub             & -       & 276              & 119          & 70:30      & Unspecified & \citetlanguageresource{addsub__hosseini2014learning} \\
      MultiArith         & -       & 420              & 180          & 70:30      & Unspecified & \citetlanguageresource{multiarith__roy2016solving}   \\
      SVAMP              & -       & 700              & 300          & 70:30      & MIT         & \citetlanguageresource{svamp__patel2021nlp}          \\
      GSM8K              & -       & 7473             & 1319         & Original   & MIT         & \citetlanguageresource{cobbe2021training}            \\
      Date Understanding & 5--6    & 258              & 111          & 70:30      & Apache-2.0  & \citetlanguageresource{big__srivastava2022beyond}    \\
      CommonSenseQA      & 5       & 9741             & 1221         & Original   & Unspecified & \citetlanguageresource{talmor2018commonsenseqa}      \\
      StrategyQA         & 2       & 1603             & 687          & 70:30      & Apache2.0   & \citetlanguageresource{strategyqa__geva2021did}      \\
      \bottomrule
    \end{tabular}
  }
  \caption{Description of datasets used in our study.}
  \label{tab:datasets}
\end{table*}

\begin{enumerate}
  \item \textbf{Supervised Fine-Tuning Model (SFT Model):} After collecting the rationale data, we fine-tune the model in the assembled high-quality rationales to get a SFT model $\mathcal{M}_{\text{sft}}$. The training objective employed for this fine-tuning remains consistent with the pre-training phase, specifically utilizing the autoregressive language modeling objective or next-token prediction \citep{gpt1__radford2018improving}. Mathematically, the objective is to minimize the language modeling loss below: \begin{equation}
          \mathcal{L} = - \sum_{t=1}^{T} \log p(x_t | x_1, \ldots, x_{t-1}; \theta)
        \end{equation} To be noticed, we use the format ``question, rationale, answer'' for training, and only calculate the loss of \textit{rationale} and \textit{answer}.

  \item \textbf{Reward Model (RM):} To train the Reward Model $\mathcal{M}{rm}$, we utilize both high-quality and low-quality rationales from the same question. The training objective for the reward model is captured by the following loss function:
        \begin{equation}
          \resizebox{\linewidth}{!}{
            $\begin{aligned}
                \mathcal{L} = & -E_{(x,y_j,y_k) \in D}[\log(\sigma(r_{\theta}(x,y_j) - r_{\theta}(x,y_k)))] 
              \end{aligned}$
          }
        \end{equation}In this function, \( r \) represents the model's score, and \( y_j \) is the preferred choice. The equation ensures that the RM assigns a higher score to the high-quality rationale \( y_j \) compared to the low-quality one \( y_k \). The second term of the loss acts as a regularizer, penalizing extreme values of the scores.

  \item \textbf{Reinforcement Learning:} Finally, we employ the fine-tuned model \( \mathcal{M}_{sft} \), and the reward model \( \mathcal{M}_{rm} \) to perform reinforcement learning in the training dataset utilizing the PPO algorithm. The SFT Model serves as a backbone, guiding the initial stages of the learning, while the RM provides the necessary feedback for refining the policy. A common issue with training the language model with RL is that the model can learn to exploit the reward model by generating complete gibberish, which causes the reward model to assign high rewards. To balance this, we add a penalty to the reward: we keep a reference of the model that we don’t train and compare the new model’s generation to the reference one by computing the KL-divergence:
        \begin{equation}
          R(x,y)=r(x,y)-\beta KL(x,y)
        \end{equation}
        where $r$ is the reward from the reward model and $KL(x,y)$ is the KL-divergence between the current policy and the reference model.
\end{enumerate}

\subsection{Strategy of Reward}

We already have a rank preference and need to design a reward strategy. We have devised three strategies to investigate the impact of the reward strategy on the model. Next, we introduce these three RL reward strategies.

\begin{itemize}
  \item \textbf{Simple RL:} During the training process, we predefined the output format, allowing us to extract the model's final answer from its output. As we train in the dataset, we can compare the model's output with the correct answer. If the output matches the correct answer, we confer a positive reward score for the output; otherwise, a negative reward score is given. This represents the simplest scenario based on our ranking preference.
  \item \textbf{Model RL:} In Simple RL, we directly compare the model's output with the correct answer. However, this approach solely discerns correct from incorrect outputs without assessing the quality of rationales that are both correct or incorrect. To address this limitation, we propose training a reward model using rationales generated from the training set, both correct and incorrect. This empowers the model to implicitly discern the quality of a rationale.
  \item \textbf{Correction RL:} We integrate the approaches of both Simple RL and the Reward Model. If the model's predicted Final Answer is incorrect, we confer a negative reward score. However, if the answer is correct, we compare the results of Simple RL with the Reward Model and allocate the greater one from the two methods. In this way, we can avoid some errors in the Reward Model in some contexts.
\end{itemize}

\section{Experiments}\label{sec:exp}

\subsection{Tasks and datasets}

Following the split of \citet{ho-etal-2023-large}, we evaluate our method in 8 datasets pertaining to three categories of complex reasoning, which are shown in \cref{tab:datasets}. These include SingleEq \cite{singleeq__koncel2015parsing}, AddSub \cite{addsub__hosseini2014learning}, MultiArith \cite{multiarith__roy2016solving}, SVAMP \cite{svamp__patel2021nlp}, GSM8K \cite{cobbe2021training}, Date Understanding \cite{big__srivastava2022beyond}, CommonSenseQA \cite{talmor2018commonsenseqa} and StrategyQA \cite{strategyqa__geva2021did}. 

\begin{table*}[ht]
  \centering
  \resizebox{\textwidth}{!}{%
    \begin{tabular}{@{}lrrrrrrrrr@{}}
      \toprule
      \multirow{2}{*}{Method}          & \multirow{2}{*}{Param} & Single         & Add            & Multi          & \multirow{2}{*}{SVAMP} & \multirow{2}{*}{GSM8K} & Date           & Common         & Strategy       \\            %
                                       &                        & Eq             & Sub            & Arith          &                        &                        & Understanding  & SenseQA        & QA             \\ \midrule   %
      \rowcolor[rgb]{0.93,0.93,0.93}\multicolumn{10}{c}{\textbf{Close-Source Models}}                                                                                                                                   \\ 
      text-davinci-003                 & 175B                   & 86.4           & 81.3           & 83.7           & 73.6                   & 59.5                   & 77.0           & 70.0           & 61.1           \\
      text-davinci-002                 & 175B                   & 82.24          & 78.99          & 78.89          & 64.67                  & 40.26                  & 73.87          & 61.75          & 53.57          \\ 
      \rowcolor[rgb]{0.93,0.93,0.93}\multicolumn{10}{c}{\textbf{Open-Source Models}}                                                                                                                                    \\ 
      StableVicuna                     & 13B                    & 62.50          & 57.14          & 43.33          & 46.67                  & 40.26                  & 45.95          & 58.64          & 41.34          \\
      LLama2-Chat                      & 7B                     & 73.03          & 68.91          & 67.22          & 53.67                  & 28.35                  & 35.14          & 56.67          & 38.14          \\
      \rowcolor[rgb]{0.93,0.93,0.93}\multicolumn{10}{c}{\textbf{Methods on Llama2 7B}}                                                                                                                                  \\ 
      Few-shot-CoT                     & 7B                     & 63.82          & 54.62          & 35.00          & 39.00                  & 14.60                  & 53.15          & 50.61          & 61.28          \\            %
      Few-shot-CoT$^{SC=8}$            & 7B                     & 67.76          & 67.23          & 55.56          & 44.67                  & 15.09                  & 35.13          & 48.40          & 62.45          \\
      Fine-tune                        & 7B                     & 71.05          & 63.87          & 11.67          & 45.67                  & 12.58                  & 64.87          & 76.58          & 65.21          \\            %
      Fine-tune-CoT (text-davinci-002) & 7B                     & 70.39          & 72.27          & 76.67          & 47.33                  & --                     & 73.88          & --             & 58.95          \\            %
      Fine-tune-CoT (STaR)             & 7B                     & 75.66          & 67.23          & 72.78          & 44.33                  & 17.29                  & 81.98          & 63.63          & 64.63          \\
      Fine-tune-CoT (Llama2)           & 7B                     & 71.05          & 65.55          & 53.33          & 40.67                  & 13.72                  & 83.78          & 69.53          & 60.84          \\            %
      Self-motivated Learning          & 7B                     & \textbf{76.32} & \textbf{76.47} & \textbf{80.00} & \textbf{55.33}         & \textbf{18.88}         & \textbf{87.39} & \textbf{77.97} & \textbf{66.08} \\            %
      \bottomrule
    \end{tabular}%
  }
  \caption{Accuracy (\%) in 8 tasks under our different models and methods. Note that the methods based on the LLama2 7B are trained in different datasets seperately.} \label{tab:results}
\end{table*}

\subsection{Comparison methods}
We present a comparison of our methods alongside several baseline methods.

\paragraph{Open/Close-Source Models}: We prompt the open-source models and the close-source models to generate the rationales and final answers.
\begin{itemize}
  \item \textbf{Open-Source Models}: It takes the instruction-tuning format of StableVicuna and LLama2-Chat with the final answer generation.
  \item \textbf{Close-Source Models}: It takes the Zero-shot-CoT format following \citet{NEURIPS2022_8bb0d291}: ``Q: <$\hat{q}_i$>. A: Let’s think step by step. <$\hat{r}^i$> Therefore, the answer is <$\hat{a}_i$>''.
\end{itemize}

\paragraph{Methods on Llama2 7b}: We compare our method with the following methods on Llama2 7B.
\begin{itemize}
  \item \textbf{Few-shot-CoT}: This method employs few-shot prompting, as outlined in \citep{wei2022chain}. And the Few-shot-CoT$^{SC=8}$ means the self-consistency and the number of samples is 8.

  \item \textbf{Fine-tune}: The model is fine-tuned in the training dataset without any rationales.

  \item \textbf{Fine-tune-CoT}: This model is fine-tuned with rationales generated by different methods.
        \begin{itemize}
          \item \textbf{Fine-tune-CoT (text-davinci-002)}: This model is fine-tuned with diverse reasoning data generated by text-davinci-002 from \citet{ho-etal-2023-large}. Due to limited training resources, \citet{ho-etal-2023-large} did not generate diverse reasoning data for all datasets (e.g. GSM8K, CommonsenseQA).
          \item \textbf{Fine-tune-CoT (STaR)}: This model is fine-tuned with the rationales generated following STaR \cite{zelikman2022star}.
          \item \textbf{Fine-tune-CoT (Llama2)}: This model is fine-tuned with the filtered rationales generated with Few-shot-CoT by Llama2 7B.
        \end{itemize}

  \item \textbf{Self-motivated Learning (Ours)}: The method, which implements reinforcement learning with PPO for the ``Fine-tune-CoT (Llama2)'' model, is described in Section \ref{sec:method}.
\end{itemize}

Due to certain policy restrictions, we cannot access OpenAI's API. Consequently, we trained the Fine-tune-CoT (text-davinci-002) model using data from \citet{ho-etal-2023-large}. Additionally, the text-davinci-002 performance results are sourced from \citet{ho-etal-2023-large}.

\subsection{Experiments Setting}

\paragraph{Implementation details.}
All experiments were conducted using the Llama2 7B model \cite{touvron2023llama} with Lora \cite{hu2021lora}.
We employ Lora to train the SFT model and Reward Model under half-precision to obtain their Lora weights $W_{SFT}$ and $W_{RM}$.
Subsequently, we use the weight $W_{SFT}$ to initialize the Lora weight of the Policy.
In Reinforcement Learning, we utilize PPO to optimize the Policy's LoRA weights $W_{Policy}$, only requiring switching between $W_{RM}$, $W_{Policy}$, and $W_{SFT}$ during the training process.
It is important to note that we did not merge the Lora weight $W_{SFT}$ with the original model during the training, ensuring that the resulting Policy weight is relatively small, which aids in memory conservation.
By adopting this approach, we can significantly save GPU memory and storage space.
The format of the training data for SFT will be formatted like the ``final answer generation'' task shown in \cref{tab:templates}.

\paragraph{Generation.}
Following \citet{wei2022chain, NEURIPS2022_8bb0d291, ho-etal-2023-large}, we employ greedy decoding to evaluate performance. For \textit{rationale generation}, we apply Few-shot-CoT with temperature sampling configured with parameters: $T = 0.8$, $TopP = 0.95$, and $Max~Length = 512$, and then use greedy decoding for \textit{final answer generation}. To optimize memory usage, we incorporate temperature sampling, but adjust the parameters to $TopP = 1.0$ and $Max~Length = 150$ during the reinforcement learning process. Detailed templates for \textit{rationale generation} and \textit{final answer generation} are shown in \cref{tab:templates}.

\paragraph{Rationale Data.} Due to resource limitation, \citet{ho-etal-2023-large} did not produce diverse reasoning for CommonsenseQA. For analogous reasons, we generated merely $5$ instances of diverse reasoning for CommonsenseQA. For mathematical problems, we randomly generated an incorrect answer between $0$ and $100$. For other datasets, we generate $8$ instances of diverse reasoning and $2$ instances of rationale for each incorrect answer and limit the rationales from text-davinci-002 to $8$ instances.

\begin{figure*}[ht]
  \centering
  \begin{subfigure}[b]{0.3\linewidth}
    \centering
    \includegraphics[width=\textwidth]{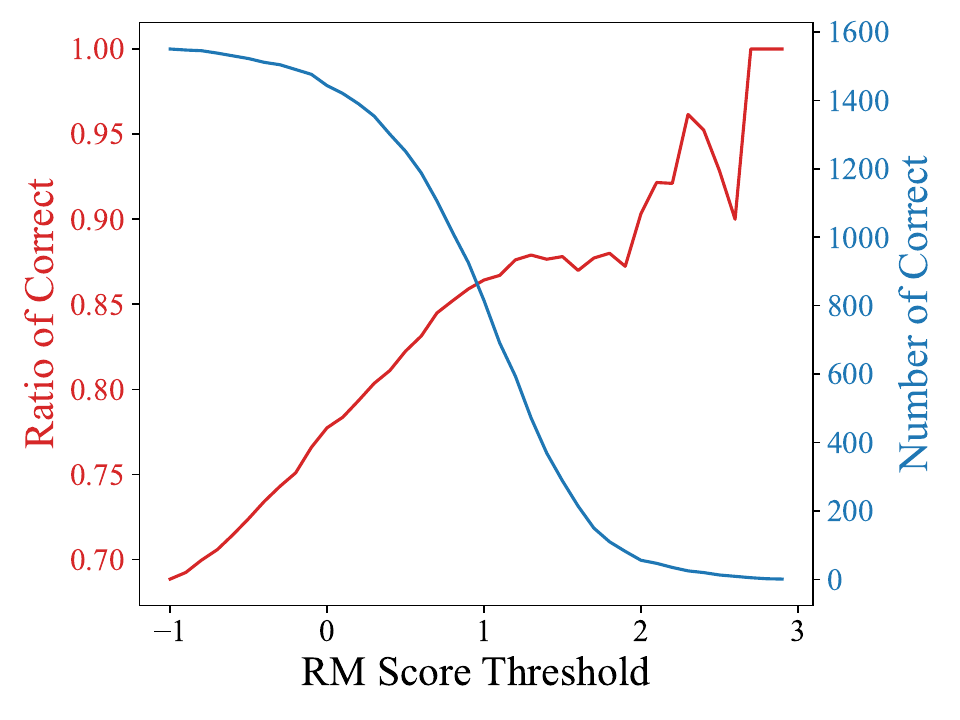}
    \caption{The ratio and count of correct reasoning vary with the reward score threshold.}
    \label{fig:rm}
  \end{subfigure}
  \hfill
  \begin{subfigure}[b]{0.3\linewidth}
    \centering
    \includegraphics[width=\textwidth]{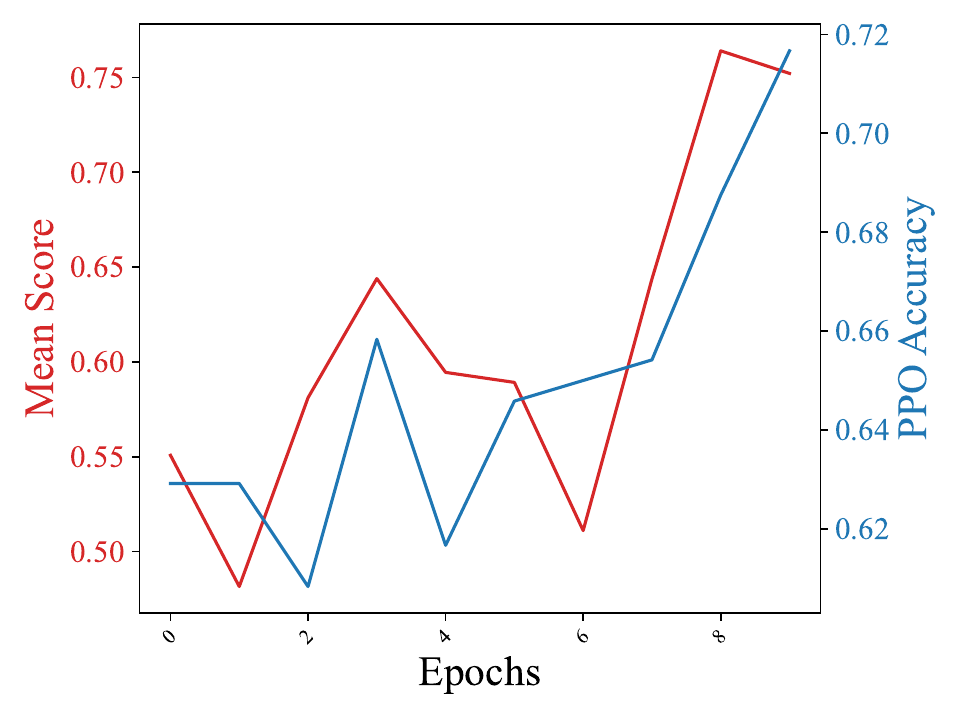}
    \caption{The variation in the accuracy of the PPO/RM model and the average score.}
    \label{fig:score}
  \end{subfigure}
  \hfill
  \begin{subfigure}[b]{0.3\linewidth}
    \centering
    \includegraphics[width=\textwidth]{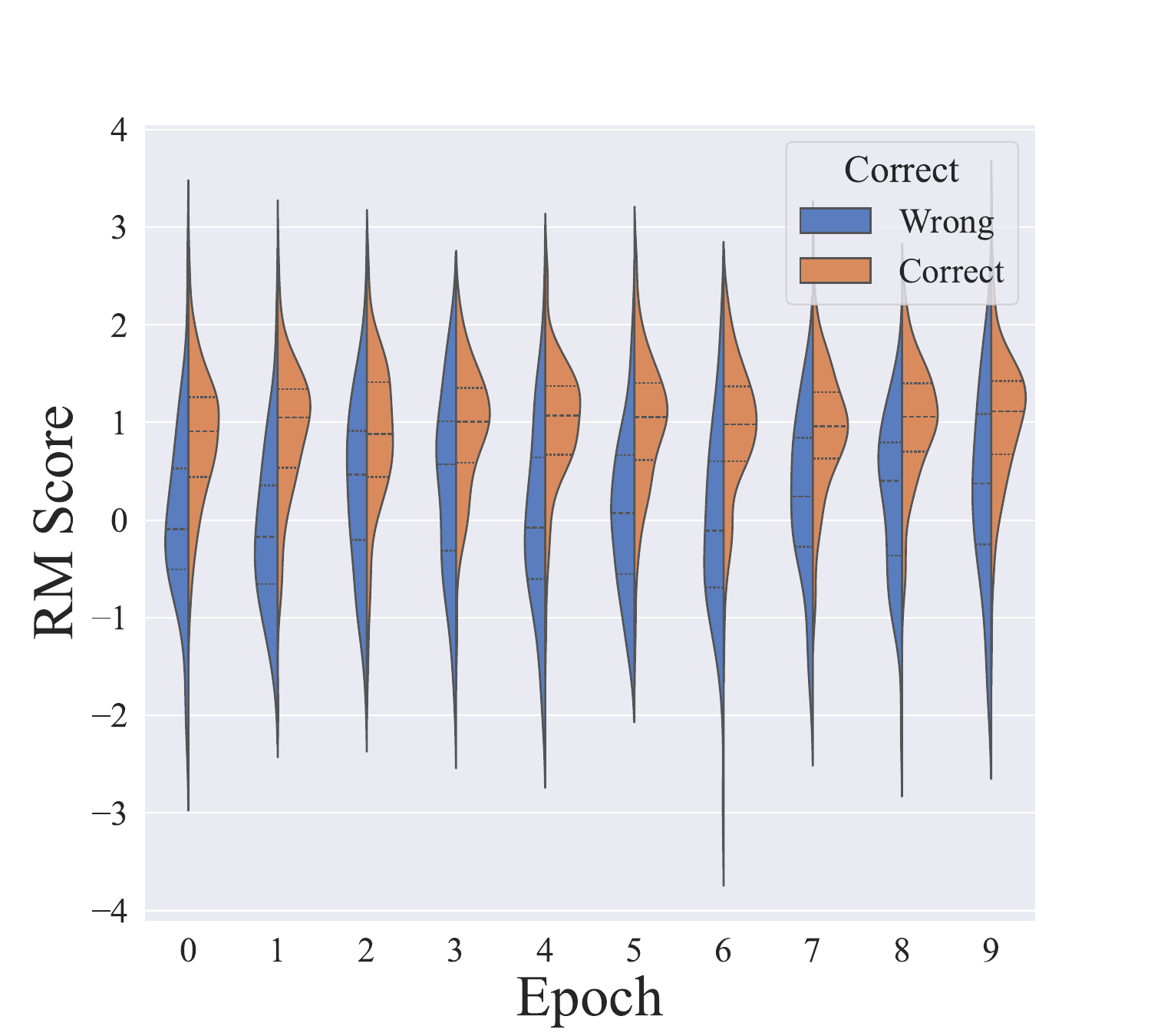}
    \caption{The score distribution of the RM model in every epoch during PPO.}
    \label{fig:box}
  \end{subfigure}
  \caption{The analysis in the SingleEq dataset. The ``RM Score Threshold'' is what we use to filter the rationales whose score is greater than the threshold during PPO. The ``PPO Accuracy'' means the accuracy in the training dataset during PPO.}
  \label{fig:analysis}
\end{figure*}

\subsection{Results}

In this section, we present the results of our experiments. We first present the results of the methods and then analyze the results of our method.

\begin{table}[t]
  \centering
  \resizebox{\linewidth}{!}{%
    \begin{tabular}{@{}lrrrr@{}}
      \toprule
      Method                           & SingleEq & AddSub & MultiArith \\ \midrule
      Fine-tune-CoT (text-davinci-002) & 70.39    & 72.27  & 76.67      \\
      \multicolumn{1}{r}{+RL}          & 71.71    & 78.16  & 80.56      \\ \midrule
      Increase                         & +1.32    & +5.89  & +3.89      \\ \bottomrule
    \end{tabular}%
  }
  \caption{The performance comparison of Fine-tune-CoT (text-davinci-002) before and after reinforcement learning (RL) implementation. The reward model used in RL is constructed based on data generated by Llama2 7B. It can be observed that the reward model trained with Llama2 7B can be effectively transferred to Fine-tune-CoT (text-davinci-002) to enhance its performance.}
  \label{tab:instruct_cot_rl}
\end{table}

\paragraph{Reasoning generation with small models is not bad.}
We employ Llama2 7B combined with Few-shot-CoT to generate rationales. After fine-tuning Llama2 7B with filtered rationales, the performance of Fine-tune-CoT surpasses the Fine-tune-CoT (text-davinci-002) in some datasets. For example, the improvements in SingleEq, Date Understanding, and StrategyQA are $0.66\%$, $9.90\%$, and $1.69\%$, respectively. This indicates that small models, when filtered, can provide valuable reasoning data. Furthermore, generating reasoning with smaller models is cost-effective.

\paragraph{Fine-tune-CoT's effectiveness is limited In commonsense reasoning.}
The gains from Fine-tune-CoT in commonsense reasoning tasks are relatively modest. It can be observed that whether using CoT generated from Llama2 or InstructionGPT, the enhancements introduced by Fine-tune-CoT are limited. Both show a performance drop compared to the direct use of Fine-tuning. In our experiments, the performance decreased by $7\%$ to $15\%$ compared to direct fine-tuning.

\paragraph{Reinforcement learning can significantly improve performance.}
The Self-motivated Learning applies PPO for reinforcement learning in the training dataset using the Fine-tune-CoT model, resulting in an average increase of $10.68\%$. This demonstrates its superior performance in various tasks, with an average accuracy of $74.22\%$. Notably, our method surpasses the performance of the model fine-tuned with CoT generated by text-davinci-002 in all datasets. Specifically, in MultiArith, Date Understanding, CommonSenseQA, and StrategyQA, our approach outperforms text-davinci-002.

\begin{table}[t]
  \centering
  \resizebox{\linewidth}{!}{%
    \begin{tabular}{@{}ccc|rrrrrrr@{}}
      \toprule
      \multicolumn{3}{c|}{State} & Single    & Add       & Multi & \multirow{2}{*}{SVAMP} & Date  & Common & Strategy                      \\
      FS                         & SFT       & RL        & Eq    & Sub                    & Arith &        & Understanding & SenseQA & QA  \\ \midrule
      \ding{51}                  & \ding{51} & \ding{51} & 83    & 5                      & 43    & 75     & 46            & 511     & 261 \\
      \ding{51}                  & \ding{55} & \ding{51} & 10    & 59                     & 13    & 19     & 7             & 46      & 45  \\
      \ding{55}                  & \ding{51} & \ding{51} & 19    & 3                      & 46    & 28     & 40            & 279     & 101 \\
      \ding{55}                  & \ding{55} & \ding{51} & 4     & 25                     & 42    & 44     & 4             & 116     & 47  \\
      \ding{51}                  & \ding{51} & \ding{55} & 2     & 0                      & 5     & 5      & 3             & 22      & 39  \\
      \ding{51}                  & \ding{55} & \ding{55} & 2     & 1                      & 2     & 18     & 3             & 39      & 76  \\
      \ding{55}                  & \ding{51} & \ding{55} & 4     & 0                      & 2     & 14     & 4             & 37      & 17  \\
      \ding{55}                  & \ding{55} & \ding{55} & 28    & 26                     & 27    & 97     & 4             & 171     & 101 \\ \bottomrule
    \end{tabular}%
  }
  \caption{We present the fluctuation in the number of samples of different states using three methods on the test set: Few-shot (FS), Supervised Fine-tuning (SFT), and Reinforcement Learning (RL). For instance, the second sequence ``\ding{51} \ding{55} \ding{51}'' indicates that the sample is correct under FS and RL, but incorrect under SFT. We can see that RL can rectify the errors introduced during SFT in the different datasets.}
  \label{tab:transfer}
\end{table}

\subsection{Analysis}

In this section, we conduct an analysis of our method using the SingleEq dataset. Our primary focus includes examining the relationship between reward scores and the quality of rationales, the correlation between given answers and rationales, the transferability of the reward model, and the impact of reinforcement learning.

\begin{table*}[ht]
  \centering
  \resizebox{\textwidth}{!}{%
    \begin{tabular}{@{}lrrrrrrrrr@{}}
      \toprule
      \multirow{2}{*}{Method} & \multirow{2}{*}{Param} & Teacher & Single         & Add            & Multi          & \multirow{2}{*}{SVAMP} & Date           & Common         & Strategy       \\            %
                              &                        & Param   & Eq             & Sub            & Arith          &                        & Understanding  & SenseQA        & QA             \\ \midrule   %
      Fine-tune-CoT           & 7B                     & 7B      & 71.05          & 65.55          & 53.33          & 40.67                  & 83.78          & 69.53          & 60.84          \\            %
      ~~+ Simple RL           & 7B                     & 7B      & 75.00          & 73.11          & 76.11          & 50.67                  & \textbf{87.39} & 77.64          & 65.21          \\            %
      ~~+ Model RL            & 7B                     & 7B      & 74.34          & 68.06          & 68.33          & 54.00                  & 86.49          & 76.33          & 64.77          \\            %
      ~~+ Correction RL       & 7B                     & 7B      & \textbf{76.32} & \textbf{76.47} & \textbf{80.00} & \textbf{55.33}         & \textbf{87.39} & \textbf{77.97} & \textbf{66.08} \\            %
      \bottomrule
    \end{tabular}%
  }
  \caption{
    Accuracy (\%) in different datasets with different Reinforcement Learning (RL) strategies. Our proposed strategy, "Correction RL," shows the highest improvement. The method ``Fine-tune-CoT + Correction RL'' is our proposed method, which is also called ``Self-motivated Learning''.
  }
  \label{tab:rl_strategy}
\end{table*}

\paragraph{The score of the reward model reflects the quality of the rationale.} As \cref{fig:rm} shows, we use the score threshold to filter the rationales during the PPO in the dataset SingleEq. We can see that the ratio of correct rationales increases with the score threshold. This means that the higher the reward model score, the more likely the rationale is correct. This shows that the reward model score can reflect the quality of the rationale to some extent.

\paragraph{The score and performance of the model are positively correlated.} As depicted in \cref{fig:score}, the average score and the performance of the model show a trend of improvement over time, although they do not always align perfectly. This misalignment might be attributed to imperfections in the reward model. \cref{fig:box} illustrates the distribution of the reward score during PPO training. It is evident that the score distribution is dispersed, particularly for incorrect rationales. This dispersion suggests that the reward model might occasionally assign high scores to incorrect rationales, reinforcing the need for our reward score corrections. Such modifications effectively address the \textit{reward hacking} challenge prevalent in RL algorithms.

\paragraph{Wrong answer leads to wrong rationales.} We use the wrong answer prompt to generate some wrong rationales, as shown in \cref{tab:prompt}. We successfully induced the model to output wrong rationales with logical errors. This may be because when the model generates rationales with in-context learning, it will try to explain the wrong label as much as possible, resulting in errors.

\paragraph{The reward model trained using this rank information exhibits a certain degree of generalization.} As depicted in \cref{tab:instruct_cot_rl}, we conducted reinforcement learning on Fine-tune-CoT (text-davinci-002) using the reward model trained with data generated by Llama 2 7B. This resulted in enhanced performance over the original model. Specifically, there was an improvement of $1.32$ in SingleEq, $5.89$ in AddSub, and $3.89$ in MultiArith, with an average enhancement of $3.70$ in performance. This indicates that the reward model trained with rank information possesses a degree of generalizability. This implies that even if we use different models to generate data for training the reward model, it can still be transferred to other models for reinforcement learning to improve their performance.

\paragraph{RL rectified some errors introduced during Supervised Fine-tuning.} We assessed the accuracy variations of samples in the test set under three methods: Few-shot (FS), Supervised Fine-tuning (SFT), and Reinforcement Learning (RL). As shown in \cref{tab:transfer}, it is evident that, compared to introducing new errors, RL generally corrects the original errors brought about by SFT. Specifically, in the AddSub dataset, RL corrected the $59$ errors that resulted from SFT without introducing any new errors. This indicates that RL can, to some extent, correct the errors introduced during SFT.

\subsubsection{Strategy of Reward}

\paragraph{Simple RL is Strong.} As \cref{tab:rl_strategy} shows, the Simple RL strategy can improve the performance of the model in all datasets. The performance of the model in the SingleEq dataset is improved by $4.95\%$, and the performance in the AddSub dataset is improved by $7.61\%$. The performance in the MultiArith dataset is improved by $22.78\%$. This shows that the Simple RL strategy can effectively improve the performance of the model. It shows that the model can learn from the correct answer and the wrong answer rank to know what rationale is correct.

\paragraph{Model RL and the Challenge of Reward Hacking.} As depicted in \cref{tab:rl_strategy}, while the model reward strategy enhances performance, it still falls short of the achievements demonstrated by the simple RL strategy. A potential explanation for this disparity is the imperfection inherent in the reward model, which can lead to \textit{reward hacking} as described by \citet{skalse2022defining}. This phenomenon may result in erroneously high scores for some rationales. A plausible cause for this could be the limited size of the training data used for the reward model. To mitigate the effects of reward hacking with limited data, we introduce the Correction RL strategy.

\paragraph{Prevent reward hacking with score correction.} Based on the two strategies, we propose a new strategy called Correction RL. Compared to the Simple RL, we use the reward model to give better fine-grained feedback to the positive example. Compared to the Model RL, we use the method of comparing the answer to avoid the error of the reward model, which significantly eliminates the impact of reward hacking. As \cref{tab:rl_strategy} shows, the Correction RL strategy can improve the performance of the model in all datasets and is superior to the other two strategies.

\section{Related Works}\label{sec:rel}

\paragraph{Reasoning Skills.}
Researchers in the literature have proposed many benchmarks requiring various reasoning skills, including
commonsense reasoning \cite{zellers-etal-2018-swag,talmor2019commonsenseqa,https://doi.org/10.48550/arxiv.1908.05739,geva2021did}
numerical reasoning \cite{dua-etal-2019-drop}, multi-hop reasoning \cite{yang-etal-2018-hotpotqa}, arithmetic reasoning \cite{koncel2015parsing,roy2015solving,miao2020diverse,patel2021nlp,cobbe2021training}, logical reasoning \cite{https://doi.org/10.48550/arxiv.2007.08124,https://doi.org/10.48550/arxiv.2002.04326}, inductive reasoning \cite{sinha2019clutrr} and tabular reasoning \cite{chen-etal-2020-hybridqa,https://doi.org/10.48550/arxiv.2105.07624}.

\paragraph{Advancements in Reasoning with Language Models.}
Language models (LMs), particularly large LMs (LLMs), have shown significant potential in addressing reasoning tasks with Chain-of-Thought \citet{wei2022chain,NEURIPS2022_8bb0d291}. This technique necessitates the model to first generate a rationale, subsequently leading to an answer. An insightful observation by \citet{wang2022self} reveals that incorporating a majority vote over multiple rationales can compensate for the shortfalls inherent in an LLM generating a singular, potentially incomplete rationale. However, the effectiveness of CoT Prompting diminishes with smaller LMs \citep{chung2022scaling}. Several studies have ventured into enhancing the reasoning abilities of LMs through diverse methodologies. For instance, \citet{deng-etal-2021-reasonbert} utilized internet-crawled data for training LMs. Techniques like logic-guided data augmentation were introduced by \citet{asai-hajishirzi-2020-logic}. On the other hand, \citet{shen2021generate,cobbe2021training,li-etal-2023-making} advocated for the training of a verifier, the task of which is to rank solutions drawn from fine-tuned LMs. An alternate approach is to endow LMs with reasoning skills by devising training samples through human-crafted templates, a method endorsed by researchers such as \citet{geva-etal-2020-injecting,yoran-etal-2022-turning,campagna-etal-2020-zero,wang-etal-2022-logic}. Taking a step further, \citet{pi2022reasoning} proposed the integration of reasoning faculties into LMs via continual pre-training on program execution data. There have been attempts to generate explanations for datasets using boosting methods by \citet{zelikman2022star} with hint. Lastly, fostering the CoT capabilities of smaller models by leveraging large models' rationale generation in diverse datasets is a concept \citet{kim2023cot,ho-etal-2023-large,wang-etal-2023-scott,li-etal-2023-symbolic}.
While many focus on enhancing the reasoning capabilities of large language models or rely on extra language models or manual efforts to generate data to improve the performance of the smaller models, our objective is to fully tap into the potential of the model itself, reduce dependency on large-scale models and manual annotations, thereby improving the reasoning prowess of these compact models.

\paragraph{Reinforcement learning from human feedback}
Reinforcement learning from human feedback (RLHF) involves training models to perform tasks through feedback obtained from human evaluators, as opposed to traditional reward signals from an environment \citep{stiennon2020learning}. Such methods have proven effective in refining models' behaviors, especially when environment rewards are sparse or ambiguous. In recent studies, such as \citet{nakano2021webgpt,ouyang2022training}, RLHF has been utilized to fine-tune large language models by collecting comparison data, where multiple model responses are ranked by quality. There are some methods do RLHF with large-scale models and extensive human feedback \citep{ouyang2022training,lightman2023lets,uesato2022solving,luo2023wizardmath}. And most methods in RLHF necessitate large models or extensive manually annotated data and typically focus on value alignment and safety alignment \cite{bai2022training,ganguli2022red,dai2023safe}. In contrast, we leverage similar techniques to enhance the model's reasoning capabilities while reducing the dependence on large models and manual annotation.

\section{Conclusion}\label{sec:conclusion}

We propose ``Self-motivated Learning'', a task-agnostic approach designed to enhance reasoning performance in LMs while decreasing reliance on large models and manual annotations. This framework is grounded in the idea that a rationale leading to the correct answer is superior to one leading to an incorrect answer. We conducted experiments across 8 datasets encompassing three categories of complex reasoning, demonstrating that our method can significantly enhance model performance without external annotation.

\section*{Acknowledgments}
We gratefully acknowledge the support of the National Natural Science Foundation of China (NSFC) via grant 62236004 and 62206078.

\section{Bibliographical References}\label{sec:reference}

\bibliographystyle{lrec-coling2024-natbib}
\bibliography{anthology,custom}

\section{Language Resource References}\label{lr:ref}
\bibliographystylelanguageresource{lrec-coling2024-natbib}
\bibliographylanguageresource{languageresource}

\end{document}